\documentclass[a4paper]{article}
\usepackage{pifont}
\usepackage{tikz}
\usetikzlibrary{decorations.pathreplacing}
\usetikzlibrary{patterns}
\usepackage{amssymb}
\usepackage{multirow}
\usepackage{cite}
\usepackage{pgfplots}
\usepackage{booktabs}
\usepackage{graphicx}
\usepackage{array,multirow}
\usepackage{INTERSPEECH2021}
\newcommand\blfootnote[1]{%
  \begingroup
  \renewcommand\thefootnote{}\footnote{#1}%
  \addtocounter{footnote}{-1}%
  \endgroup
}
\newcommand{\cmark}{\ding{51}}%
\newcommand{\xmark}{\ding{55}}%
\newcommand{\quotes}[1]{``#1''}

\title{Building an ASR Error Robust Spoken Virtual Patient System in a Highly Class-Imbalanced Scenario Without Speech Data}
\name{Vishal Sunder$^*$, Prashant Serai$^*$, Eric Fosler-Lussier}
\address{
  The Ohio State University
 }
\email{sunder.9@osu.edu, serai.1@osu.edu, fosler@cse.ohio-state.edu}

\begin{document}

\maketitle
\begin{abstract}
A Virtual Patient (VP) is a powerful tool for training medical students to take patient histories, where responding to a diverse set of spoken questions is essential to simulate natural conversations with a student. The performance of such a Spoken Language Understanding system (SLU) can be adversely affected by both the presence of Automatic Speech Recognition (ASR) errors in the test data and a high degree of class imbalance in the SLU training data. While these two issues have been addressed separately in prior work,  we develop a novel two-step training methodology that tackles both these issues effectively in a single dialog agent. As it is difficult to collect spoken data from users without a functioning SLU system, our method does not rely on spoken data for training, rather we use an ASR error predictor to \quotes{speechify} the text data. Our method shows significant improvements over strong baselines on the VP intent classification task at various word error rate settings. 

\end{abstract}

\section{Introduction}
\label{sec:intro}
\blfootnote{*equal contribution}

The Virtual Patient (VP) Spoken Dialog System (SDS) is a special-purpose conversation agent developed at The Ohio State University to train medical students to take patient histories \cite{danforth2009development,danforth2013can,maicher2019using}. VP is a simple question classification model where a student question is mapped into one of several canonical questions and a deterministic response is then generated. Like many traditional SDS, VP uses an upstream Automatic Speech Recognizer (ASR) and a downstream Spoken Language Understanding system (SLU) trained to understand text input. But the SLU component for such a system poses two main challenges.

\begin{figure}[h]
    \hfill
    \centering
    \centerline{\includegraphics[width=0.6\columnwidth]{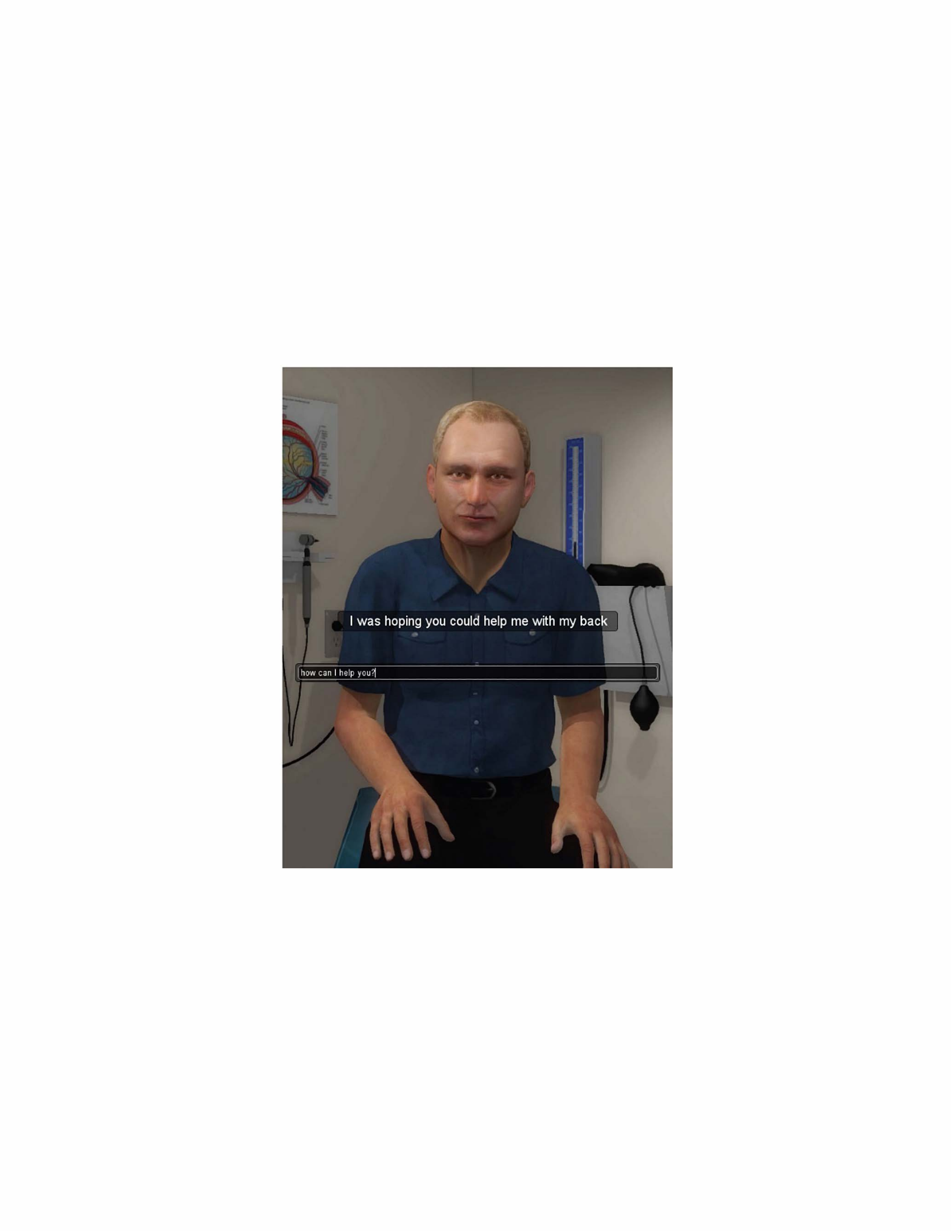}}
\caption{Interface of OSU's Virtual Patient.}
\label{fig:avatar}
\end{figure}

The first challenge is the high degree of class-imbalance in the intent classification data. The VP dataset has 376 classes spanning over a training size of 11,584. This is done to cater to fine-grained user queries, ranging from the most frequent \quotes{How are you today?} to the most infrequent \quotes{Are you nervous?}. Given the small size of this dataset, it becomes difficult for a model to deal with the sparsity in class distribution.

The second challenge is to make the SLU component robust to ASR errors. A possible strategy is to collect speech data and use the ASR hypotheses for training the SLU \cite{ruan2020towards}, but it is not straightforward to collect speech data without a usable speech based VP. In the beginning, it is simpler to build a rule-based dialog agent like ChatScript \cite{wilcox2011chatscript} to interact with the user and collect typed data \cite{danforth2009development, danforth2013can}. Once enough data is collected, a sophisticated neural model can be trained which is used to collect more data, making this a bootstrapping process. It is only after a reasonable SLU system can be deployed, that collection of spoken data from users becomes feasible. This results in a dataset with either zero or only limited speech.

\begin{figure*}
    \centering
    \begin{tikzpicture}
        \node[text width=4cm, align=left] at (-2,2) {\large \underline{\textbf{Pretraining}}};
        
        \node[text width=2cm, align=left] at (0,1.0) {$x_i$};
        \node[text width=2cm, align=left] at (0,0.2) {$x_j$};
        \draw[-stealth] (-0.6,1.0) -- (0,1.0);
        \draw[-stealth] (-0.6,0.2) -- (0,0.2);
        \filldraw[fill=white, draw=black] (0,0) rectangle (2.4,1.2);
        \node[text width=2.2cm, align=center] at (1.2,0.6) {\footnotesize Pretrained ASR Error Predictor};
        
        \draw[decoration={brace,raise=5pt,amplitude=5pt},decorate] (0,1.2) --  (2.4,1.2);
        \node[text width=2.5cm, align=center] at (1.2,2.4) {\footnotesize Error hallucination with $\epsilon$ probability (helps with ASR errors)};
        
        \filldraw[fill=lightgray, draw=black] (3,0) rectangle (5.4,1.2);
        \draw[-stealth] (2.4,1.0) -- (3.0,1.0);
        \draw[-stealth] (2.4,0.2) -- (3.0,0.2);
        \node[text width=2.2cm, align=center] at (4.2,0.6) {Encoder ($\boldsymbol \theta$)};
        \draw[stealth-, dashed] (3.0,1.4) -- (5.4,1.4);
        \node[text width=2cm, align=center] at (4.2,1.7) {$\frac{\partial \mathcal{L}_{con}}{\partial \boldsymbol \theta} + \frac{\partial \mathcal{L}_{mix}}{\partial \boldsymbol \theta}$};
        
        \draw[-stealth] (5.4,1.0) -- (6.0,0.6);
        \node[text width=1.0cm, align=center] at (5.8,1) {\scriptsize $\textbf{r}_i$};
        \draw[-stealth] (5.4,0.2) -- (6.0,0.6);
        \node[text width=1.0cm, align=center] at (5.8,0.2) {\scriptsize $\textbf{r}_j$};
        
        \filldraw[fill=white, draw=black] (6.6,0.6) circle[x radius=0.6, y radius=0.2];
        \node[text width=1.2cm, align=center] at (6.6,0.6) {\footnotesize mixup};
        \draw[-stealth] (7.2,0.6) -- (8.4,0.6);
        \node[text width=1.5cm, align=center] at (7.7,1.01) {\scriptsize $\lambda \textbf{r}_i + (1-\lambda) \textbf{r}_j$};
        
        \filldraw[fill=lightgray, draw=black] (8.4,0) rectangle (10.8,1.2);
        \node[text width=2.2cm, align=center] at (9.6,0.6) {Classifier ($\boldsymbol \phi$)};
        \draw[stealth-, dashed] (8.4,1.4) -- (10.8,1.4);
        \node[text width=2cm, align=center] at (9.6,1.7) {$\frac{\partial \mathcal{L}_{mix}}{\partial \boldsymbol \phi}$};
        
        \draw[decoration={brace,raise=5pt,amplitude=5pt},decorate] (3,1.8) --  (10.8,1.8);
        \node[text width=7.0cm, align=center] at (6.9,2.4) {\footnotesize Pairwise training (helps with class imbalance)};
        
        \draw[dotted] (-4,-0.25) -- (13, -0.25);
        \node[text width=4cm, align=left] at (-2,-1) {\large \underline{\textbf{Fine tuning}}};
        
        \node[text width=3cm, align=left] at (1.6,-2.0) {$x_i$ \footnotesize (Clean text)};
        \node[text width=3cm, align=left] at (1.6,-2.8) {$\widetilde{x_i}$ \footnotesize (Error text)};
        \draw[-stealth] (2.1,-2.0) -- (2.7,-2.0);
        \draw[-stealth] (2.1,-2.8) -- (2.7,-2.8);
        \filldraw[fill=lightgray, draw=black] (2.7,-3) rectangle (5.1,-1.8);
        \node[text width=2.2cm, align=center] at (3.9,-2.4) {Encoder ($\boldsymbol \theta$)};
        \draw[stealth-, dashed] (2.7,-1.6) -- (5.1,-1.6);
        \node[text width=1.7cm, align=center] at (3.9,-1.3) {$\frac{\partial \mathcal{L}_2}{\partial \boldsymbol \theta}$};
        
        \draw[-stealth] (5.1,-2.0) -- (5.7,-2.0);
        \draw[-stealth] (5.1,-2.8) -- (5.7,-2.8);
        \filldraw[fill=lightgray, draw=black] (5.7,-3) rectangle (8.1,-1.8);
        \node[text width=2.2cm, align=center] at (6.9,-2.4) {Classifier ($\boldsymbol \phi$)};
        \draw[stealth-, dashed] (5.7,-1.6) -- (8.1,-1.6);
        \node[text width=1.7cm, align=center] at (6.9,-1.3) {$\frac{\partial \mathcal{L}_2}{\partial \boldsymbol \phi}$};
        
        \draw[decoration={brace,raise=5pt,amplitude=5pt},decorate] (2.7,-1.2) --  (8.1,-1.2);
        \node[text width=5.4cm, align=center] at (5.4,-0.6) {\footnotesize Fine tuning with $\mathcal{L}_2$ (helps with ASR errors)};

    \end{tikzpicture}
    \caption{Model overview. Solid arrows indicate forward propagation. Dashed arrows indicate backpropagation. The grey rectangles contain trainable parameters. The white rectangle has frozen parameters.}
    \label{fig:model_over}
\end{figure*}
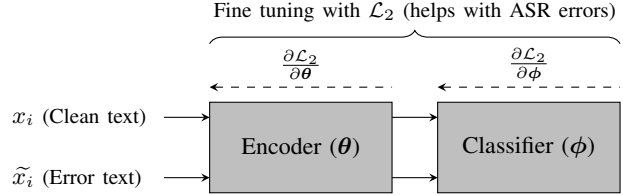



Therefore, to build a robust dialog agent in a scenario like VP, we argue that the above challenges need to be addressed together. Prior studies have dealt with these two issues separately. To deal with class-imbalance in VP data, Jin et al. \cite{jin2017combining} combine a rule-based system with a neural model and show that the rule-based system helps improve performance on rare classes. Recently, a self-attention based RNN architecture~\cite{stiff2020self} combined with a pairwise training approach~\cite{sunder2020handling} has shown significant improvements over previous work in class-imbalanced settings.


To build ASR an error robust SLU system, Ruan et al. \cite{ruan2020towards} have proposed a loss function that uses both the ASR hypothesis and the gold text, but this technique assumes the availability of parallel speech and text data. In the VP domain, text-to-phonetic data augmentation has shown good performance on ASR transcripts \cite{stiff2019improving}. Simulating ASR errors has also been shown to improve robustness in SLU \cite{serai2022hallucination, namazifar2020warped}.



We show that although previously proposed methods for building ASR error robustness give some performance boost in the VP domain, this improvement is constrained by a high degree of class imbalance. This work is a first step in building a single model that deals with these two issues in a unified way. In particular, we do not assume access to any speech data during training and utilize the recently proposed ASR error predictor \cite{serai2022hallucination} to simulate ASR errors and use these errors to train a neural model in a pairwise fashion as done by Sunder et al. \cite{sunder2020handling}. The pairwise training helps with the class imbalance issue. We also introduce a novel fine-tuning step which uses the simulated ASR errors to reduce the gap between model performance on clean text and ASR hypothesis. Thus, our approach is a two-step training procedure that is designed so that the SLU model is robust to class-imbalance and ASR errors both. The proposed methodology outperforms all baselines in two different scenarios, when no speech data is available and when only a limited amount of speech data is available during training. 


\section{ASR Error Predictor}
\label{sec:asr_err_pred}
We use the recently proposed joint phonetic and word level error predictor model~\cite{serai2022hallucination} for the purpose of translating true text (gold transcripts free from ASR errors) to recognized text (transcription hypotheses with hallucinated ASR errors). The architecture is based on a convolutional sequence to sequence framework~\cite{gehring2017convolutional} and consists of two encoders and one decoder. The first encoder takes a word sequence input corresponding to the true text, the second encoder takes the phoneme sequence input corresponding to the same, and the decoder produces predictions for the recognized word sequence. The encoders consist of 6 residual CNN layers each, and the decoder consists of 3 residual CNN layers that can each attend to both the encoders together. In order to encourage the decoder layers to attend to both encoders, an ``encoder dropout'' mechanism is used at train time where we randomly drop one of the two encoders with a certain probability during training. We pretrain the ASR error predictor and freeze it's weights, not allowing the loss from the SLU model to backpropagate through it.

\section{SLU Model}
\label{sec:utt_class}
The SLU model consists of an encoder and a classifier. The encoder is a bidirectional GRU \cite{cho2014learning} with a self-attention mechanism \cite{lin2017structured} which converts the GloVe embedding matrix \cite{pennington2014glove} of an utterance text to a vector representation. This vector is fed to a 3-layer fully connected classifier MLP with $tanh$ activations.


To deal with class-imbalance in the VP dataset, we use the recently proposed contrastive loss and interpolation based pairwise training framework \cite{sunder2020handling} to train the above model. We also utilize the pretrained ASR error predictor to inject errors into the training input for the SLU model (section \ref{subsec:hall_err}). Further, fine tuning of the above trained model is done to bring the predictions on the ASR text closer to the corresponding clean text (section \ref{subsec:finetune}). The overview of our model is given in figure \ref{fig:model_over}.

\subsection{Pairwise Training Framework}
\label{subsec:pair_fram}
Given the dataset $\{(x_i,y_i)\}_{i=1}^n$, where $x_i$ is an utterance text and $y_i$ its label, pairs of training instances are created. Each such instance is denoted as $(x_i,x_j,y_i,y_j)$. The encoder converts the GloVe representation of $x_i$ to $\textbf{r}_i$ and $x_j$ to $\textbf{r}_j$. Then, a contrastive loss is computed as,
\begin{equation*} \label{eq:l_con}
    \begin{split}
        \mathcal{L}_{con} &= \frac{1}{2}y_{pair}\{max(0,D-m_{pos})\}^2 \\&+ \frac{1}{2}(1-y_{pair})\{max(0,m_{neg}-D)\}^2
    \end{split}
\end{equation*}

Here, $D$ is the euclidean distance between $\frac{\textbf{r}_i}{\left\Vert \textbf{r}_i \right\Vert_2}$ and $\frac{\textbf{r}_j}{\left\Vert \textbf{r}_j \right\Vert_2}$; $y_{pair} = 1$ if $y_i = y_j$ and $0$ otherwise. $m_{pos}$ and $m_{neg}$ are positive and negative margins set to $0.8$ and $1.2$ respectively.

The contrastive loss only trains the encoder. To train the classifier, the \textit{mixup} strategy is used. A mixed representation of a paired training instance is created by interpolating between their encoded representation $\textbf{r}_i$ and $\textbf{r}_j$. The same is done to the one-hot representation of the labels ($\textbf{y}_i$ and $\textbf{y}_j$). This creates a new training instance $(\textbf{r}_{mix}, \textbf{y}_{mix})$ as,
\begin{equation*} \label{eq:mix}
    \begin{split}
        \textbf{r}_{mix} &= \lambda \textbf{r}_i + (1-\lambda) \textbf{r}_j \\
        \textbf{y}_{mix} &= \lambda \textbf{y}_i + (1-\lambda) \textbf{y}_j
    \end{split}
\end{equation*}
$\lambda$ is sampled from 
$\lambda \sim \text{Beta}(\alpha, \alpha)$, where $\alpha$ is a hyperparameter. The representation $\textbf{r}_{mix}$ is fed to the classifier to get an output distribution $p(y|x_{i},x_{j})$. A mixup loss is computed as a KL-divergence term,
\begin{equation*} \label{eq:l_mix}
    \begin{split}
        \mathcal{L}_{mix} = KL(\textbf{y}_{mix}, p(y|x_{i},x_{j}))
    \end{split}
\end{equation*}
The final loss for training the utterance classification model is,
\begin{equation*} \label{eq:l_1}
    \begin{split}
        \mathcal{L}_{1} = \beta \mathcal{L}_{con} + (1-\beta) \mathcal{L}_{mix}
    \end{split}
\end{equation*}
Here, $\beta \in [0, 1]$ is tuned on the development set.




\subsection{Hallucination of ASR Errors}
\label{subsec:hall_err}
The pairwise training framework helps to deal with the high class imbalance issue effectively. But during inference, we also expect performance degradation due to the presence of ASR errors. To tackle this issue, we utilize the pretrained ASR error predictor. During pairwise training, for a paired training instance $(x_i,x_j)$, we sample pseudo speech recognized alternatives for $x_i$ and $x_j$ using the ASR error predictor. These alternatives are used as the replacement for the original training example with a probability $\epsilon$ (a hyperparameter). This strategy has shown effectiveness in dealing with ASR errors in downstream tasks \cite{stiff2019improving,serai2022hallucination} and is called the \quotes{hallucination} strategy.

\subsection{Fine Tuning}
\label{subsec:finetune}
After the utterance classification model is trained using the above methods, we try to bring the model predictions on errorful text and gold text closer by using a loss function proposed by Ruan et al. \cite{ruan2020towards}. We compute the cross-entropy losses on clean text ($x_i$) and a corresponding errorful text ($\widetilde{x_i}$). The  errorful text ($\widetilde{x_i}$) is the ASR hypothesis when available and sampled from the ASR error predictor otherwise. We also add to this the KL-divergence loss on model predictions for both. Formally,
\begin{equation*} \label{eq:l_2}
    \begin{split}
        \mathcal{L}_{2} = CE(x_i,y_i) + \eta(CE(\widetilde{x_i},y_i) + KL(p(y|x_i), p(y|\widetilde{x_i}))
    \end{split}
\end{equation*}
Here, $CE(.,y_i) = -\log p(y_i|.)$ and $\eta \in \{0.1, 1.0, 10.0\}$ is a hyperparameter tuned on the development set.

Unlike previous work, we use this only as a fine-tuning criterion and do not assume access to real ASR transcripts.

\subsection{Inference}
\label{subsec:infer}
At test time, we combine a 1-nearest-neighbor search with the classifier's prediction \cite{sunder2020handling}. We cache the encoder representations for all utterances in the training set along with their 4 closest ASR alternatives in the euclidean space from the error predictor. 

Given a test instance $x_{test}$, we first encode it using the encoder and then perform a 1-nearest-neighbor search\footnote{We use the FAISS toolkit for efficient search \cite{johnson2019billion} (https://github.com/facebookresearch/faiss)} on the cached training set. Thus, we get a closest distance score for each class. We convert this into a class distribution $p_{nn}(y|x_{test})$ by first inverting the distance scores, z-score normalization and then taking a softmax. The pre-softmax output of the classifier is also z-score normalized before passing to the softmax which gives the class distribution $p(y|x_{test})$. The final prediction is computed as,
\begin{equation*} \label{eq:p_final}
    \begin{split}
        p_{final}(y|x_{test}) &= \gamma p_{nn}(y|x_{test}) + (1 - \gamma)p(y|x_{test})
    \end{split}
\end{equation*}
$\gamma \in [0,1]$ is tuned on the development set. 

\section{Experimental setup}
\label{sec:exp_set}

\begin{table*}
    \centering
    \caption{Accuracy/Macro-F1 scores using different training methods. Italic numbers indicate the oracle performance, i.e. when no errors are present. Bold numbers indicate best performance in the corresponding column; classification is either with softmax classifier ($^*$) or combined softmax/nearest neighbor ($^\dag$). (2), (3) were proposed specifically to help with ASR errors. (4) was proposed to help with class-imbalance. (5), (6) and (7) are variants of our proposal that help with both these issues. Results averaged over 10 runs.}
    \resizebox{\textwidth}{!}{\begin{tabular}{@{}llccccccccccc@{}}\toprule
        \multicolumn{2}{l}{WER in test set} && 0\% && \multicolumn{2}{c}{8.9\% (ASR system 1)} && \multicolumn{2}{c}{12.5\% (ASR system 2)} &&  \multicolumn{2}{c}{41.1\% (ASR system 3)} \\
        \cmidrule{4-4} \cmidrule{6-7} \cmidrule{9-10} \cmidrule{12-13} 
        \multicolumn{2}{l}{Some real data from ASR system 1 in train set?} && \xmark & & \xmark & \cmark & & \xmark & \cmark & & \xmark & \cmark \\ 
        \midrule
        Helps with & Training & \multicolumn{11}{c}{Accuracy/Macro-F1 (in \%)} \\
        \midrule
        None (baseline) & (1)$^*$ cross-entropy \cite{stiff2020self} && \textit{79.7/59.8} && 78.0/57.8 & 79.3/58.7 && 76.1/56.6 & 77.0/57.4 && 65.6/45.8 & 66.7/47.3 \\
        \midrule
        \multirow{2}{*}{ASR Errors} & (2)$^*$ cross-entropy w/ hallucination \cite{serai2022hallucination} && - && 78.3/58.3 & 79.5/59.0 && 77.4/58.0 & 77.8/57.8 && 67.9/48.6 & 67.5/48.5 \\
        & (3)$^*$ $\mathcal{L}_{2}$ only \cite{ruan2020towards} && - && 78.6/58.6 & 79.4/59.8 && 77.5/58.0 & 77.4/58.5 && 67.9/48.8 & 67.2/47.7 \\
        \midrule
        Class-Imbalance & (4)$^\dag$ pairwise only \cite{sunder2020handling} && \textit{80.6/65.7} && 79.0/63.7 & 79.7/64.5 && 77.4/61.7 & 77.3/62.0 && 66.9/50.4 & 67.0/50.7 \\
        \midrule
        \multirow{3}{*}{Both} & (5)$^\dag$ pairwise w/ hallucination && - && 79.3/64.3 & 79.8/64.8 && 77.9/62.7 & 77.6/62.2 && 67.9/52.0 & 67.1/51.2 \\
        & (6)$^\dag$ pairwise + fine tune with $\mathcal{L}_{2}$ && - && 79.8/64.5 & 80.2/64.6 && 77.9/62.2 & 77.9/62.3 && 67.6/50.9 & 68.3/51.6 \\
        & (7)$^\dag$ pairwise w/ hallucination + fine tune with $\mathcal{L}_{2}$ && - && \textbf{79.9/65.1} & \textbf{80.4/65.1} && \textbf{78.4/63.1} & \textbf{78.3/62.8} && \textbf{68.5/52.4} & \textbf{68.5/52.0} \\
        \bottomrule
    \end{tabular}}
    \label{tab:results}
\end{table*}

\subsection{Virtual Patient Dialog Data}
\label{subsec:data}

\begin{figure}[h]
    \caption{Distribution of classes in the training set of VP. Over $70\%$ of the data belongs to the first quintile of classes}
    \label{fig:cls_dist}
    \centering
    \centerline{\includegraphics[width=0.8\columnwidth]{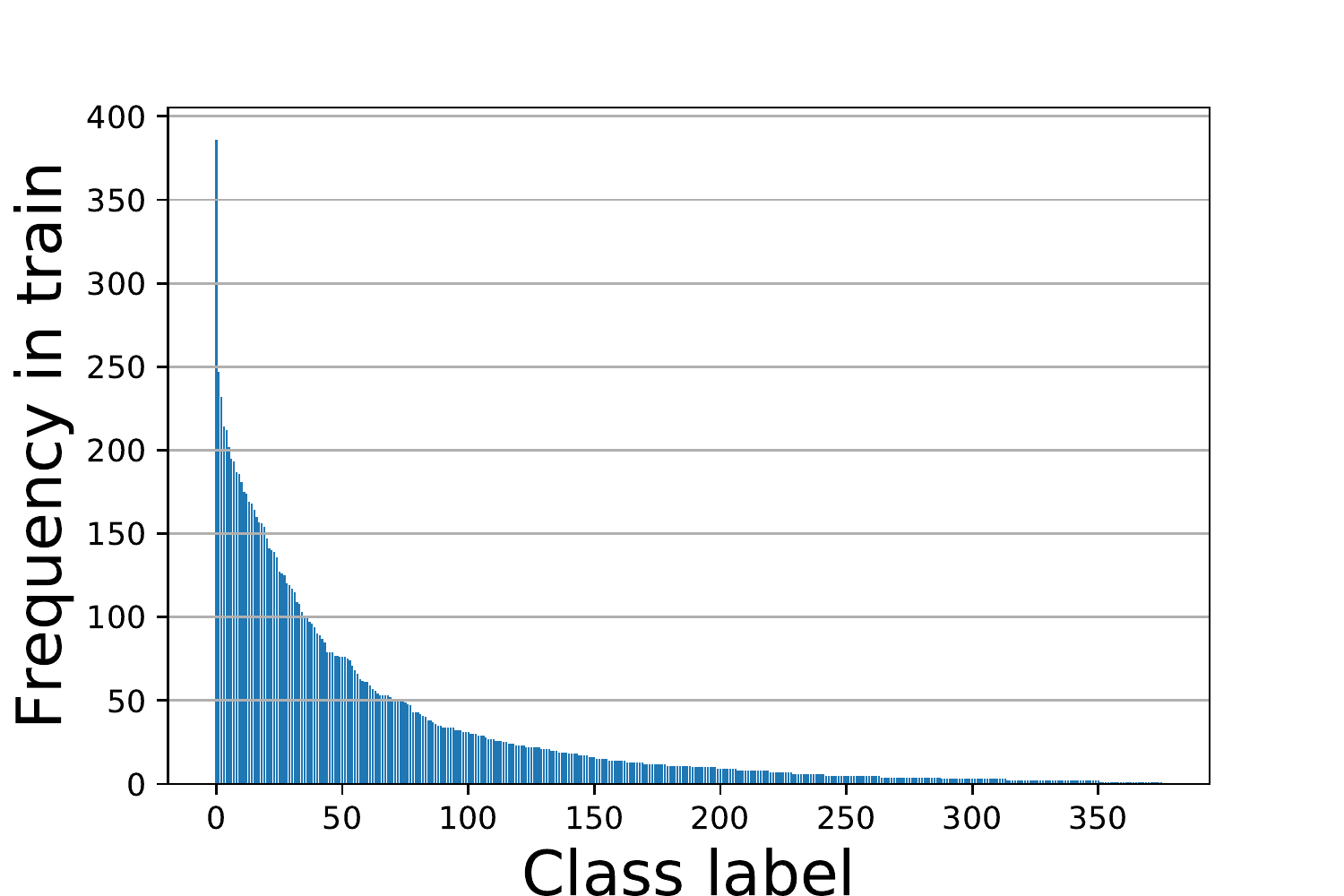}}

\end{figure}

The Virtual Patient data is a dataset of conversations between a medical student and a patient (a virtual agent) experiencing back pain. Each student utterance is classified to one of 376 possible classes and each class has a fixed agent response. The total size of the train, dev and test set is 11584, 1635 and 5016 student utterances respectively. This is a highly class imbalanced dataset as evident from the data distribution in figure \ref{fig:cls_dist}.

The data was collected in a bootstrapping manner wherein some initial (typed) conversations were collected using a rule based agent (ChatScript \cite{wilcox2011chatscript}) and after a neural model was trained using this data, spoken data was collected by deploying this model. Thus, 6711 utterances in the training set were typed and the rest were spoken. The spoken utterances were transcribed manually and by a cloud based ASR system. We make sure that all the data in the dev and test set is spoken data.


\subsection{Data for Training the ASR Error Predictor}
Our cloud based ASR system used at test-time didn't offer publicly available error characterization data, and we had cost and rate limit constraints on how much data we could collect, so for our training set we used characterization data available from an unrelated but homegrown ASR system's transcriptions of the Fisher corpus used for a prior version of the error predictor\cite{serai2019improving}. This resulted in a set of 1.8 million odd pairs of true text and recognized text used for training the error predictor. In order to finetune our errors to our test-time scenario, we used a randomly selected subset of 100k utterances from the Fisher corpus that we passed through the cloud based ASR system. In the experiments where some real ASR data from the Virtual Patient project was assumed available, the pairs of manual transcripts and cloud ASR transcripts for the spoken portion of the train set were added to the finetuning subset.

\subsection{Training details}
For the error predictor, we train with a Nesterov accelerated gradient descent optimizer~\cite{sutskever2013importance} for 60 epochs with a learning rate of 0.1 and a momentum of 0.99 on the training data from the homegrown ASR system, followed by 15 additional epochs on the finetuning data from the cloud based ASR system.

We train the SLU model using early stopping with a patience of 6 epochs. We use the Adam optimizer \cite{kingma2014adam} with learning rate of $1e-3$ for pairwise training and $8e-5$ for fine-tuning. We add a dropout of 50\% for every layer in the neural network.

\section{Results and Discussion}
\label{sec:result}


\begin{table}[]
    \centering
    \caption{Ablation study on test set with WER of 8.9\%. $\eta$ refers to the hyperparameter in $\mathcal{L}_{2}$. Bold represents best performance.}
    \resizebox{\columnwidth}{!}{
    \begin{tabular}{@{}lcc@{}}\toprule
        Training method & $\eta \neq 0$ & $\eta = 0$ \\
        \midrule
        pairwise + fine tune with $\mathcal{L}_2$ & \textbf{79.8}/\textbf{64.5} & 79.7/64.2 \\
         pairwise w/ hallucination + fine tune with $\mathcal{L}_2$ & \textbf{79.9}/\textbf{65.1} & 79.8/64.7  \\
        \bottomrule 
    \end{tabular}}
    \label{tab:ablation}
\end{table}

The results are shown in table \ref{tab:results}. To test the effectiveness of our model in noisy acoustic conditions or a possible domain mismatch for an off-the-shelf ASR~\cite{mani2020asr}, we show results when the test data has errors from 3 different ASR systems. Training is done in two scenarios: a) no real ASR hypotheses is used, i.e. all utterances in train were either typed text or manual transcriptions; b) the available ASR hypotheses is present in train. When present, these hypotheses come from a fixed ASR (system 1). This is a deployment setting where it is not feasible to re-train the SLU with new hypotheses if the upstream ASR changes. We use 4 prior studies as our baselines:

\noindent \textbf{(1) cross-entropy} \cite{stiff2020self} A self-attention GRU trained with cross-entropy loss for classification.

\noindent \textbf{(2) cross-entropy w/ hallucination} \cite{serai2022hallucination} Same as 1. except the input training instance is randomly replaced (with probability $\epsilon$) by an errorful alternative sampled from the error predictor.

\noindent \textbf{(3) $\mathcal{L}_{2}$ only} \cite{ruan2020towards} The fine-tuning objective in section \ref{subsec:finetune} is used as the only criterion without any pretraining step.

\noindent \textbf{(4) pairwise only} \cite{sunder2020handling} The pairwise training described in section \ref{subsec:pair_fram} is used without any hallucination or fine-tuning.

The last 3 rows of table \ref{tab:results} show the performances our proposed training methods.

\noindent \textbf{(5) pairwise w/ hallucination} Pairwise training is used along with error hallucination. No fine-tuning is done.

\noindent \textbf{(6) pairwise + fine tune with $\mathcal{L}_{2}$} Pairwise training is used without error hallucination but fine tuning with $\mathcal{L}_2$ is done.

\noindent \textbf{(7) pairwise w/ hallucination + fine tune with $\mathcal{L}_{2}$} All the components are combined.

Note that previous work to handle ASR errors improve the accuracy and macro-F1 scores over the baseline ((1) $\rightarrow$ (2) and (1) $\rightarrow$ (3)). However, pairwise training alone shows significant improvements of 5.9\% and 5.8\%  in macro-F1 score with 8.9\% WER without and with real ASR data respectively ((1) $\rightarrow$ (4)). A similar level of improvement is seen in other settings as well. As the pairwise training framework is designed specifically to deal with class-imbalance, such high improvements in F1 show that this is a significant problem in the dataset. This shows that the possible improvements from (1) $\rightarrow$ (2) and (1) $\rightarrow$ (3) are constrained by class-imbalance and thus validate our proposal to tackle these two issues simultaneously.

Our proposed training methods which combine the benefits of the above baselines lead to further improvements. Notice that pairwise training with hallucination (row (5)) improves performance in terms of both accuracy and F1 score over just pairwise training (row (4)). We see a mean improvement of 0.4\% in accuracy and 0.7\% in F1 score from (4) $\rightarrow$ (5) without any real ASR data across different settings. This shows that the hallucinated errors from the ASR error predictor help pairwise training to additionally build robustness against ASR errors.

We now fine tune a model with $\mathcal{L}_2$, pretrained using pairwise training without hallucination. This gives us a mean improvement of 0.7\% in accuracy and 0.5\% in F1-score ((4) $\rightarrow$ (6)). This validates the fine tuning criterion.

We see the best performance across all settings when we combine all the techniques (row (7)). This improvement is on accuracy and F1 both, which shows that we have successfully combined the ability of pairwise training to handle class imbalance with effective strategies to deal with ASR errors.

Note that our technique leads to significant improvements even when the off-the-shelf ASR system does not perform well. With higher WERs, the performance is better without any real ASR data than with it. This may be because the real ASR data present in the training set does not have the same type of errors seen at test time. This shows the need for SLU models to not rely on static ASR data during training. Our approach of using an error predictor model is a step in this direction.

\noindent \textbf{Ablation Study:} To assess the effect of fine-tuning with ASR errors, we perform an ablation study by setting $\eta = 0$ in $\mathcal{L}_{2}$ (table \ref{tab:ablation}). This study is performed in the case when there are no real ASR hypothesis in the training set and the test set has a WER of 8.9\%. The fact that we notice a performance degradation when $\eta = 0$ indicates that the errors generated by the ASR error predictor carry useful information about the real-world error. However, we believe that there is scope for more improvements in the fine-tuning stage by exploring better ways to distill more knowledge from clean text to ASR text \cite{kim2020two}.

\section{Conclusion}

In this paper, we explored methods to tackle the problem of class-imbalance and ASR errors together for SLU in a special-purpose, low-resource SDS. We introduced a training method that effectively combines a pairwise training framework with a simple fine-tuning criteria to achieve significant improvements over strong baselines. Our technique assumes that there are no real ASR transcripts available during training and thus uses a pretrained ASR error predictor to simulate ASR errors. Future work should explore ways to distill knowledge from clean text to ASR text directly during pairwise training.

\section{Acknowledgements}

This material builds upon work supported by the National Science Foundation under Grant No. 1618336.  We gratefully acknowledge the OSU Virtual Patient team for their assistance.

\bibliographystyle{IEEEtran}

\bibliography{mybib}

\begin{thebibliography}{10}
\providecommand{\url}[1]{#1}
\csname url@samestyle\endcsname
\providecommand{\newblock}{\relax}
\providecommand{\bibinfo}[2]{#2}
\providecommand{\BIBentrySTDinterwordspacing}{\spaceskip=0pt\relax}
\providecommand{\BIBentryALTinterwordstretchfactor}{4}
\providecommand{\BIBentryALTinterwordspacing}{\spaceskip=\fontdimen2\font plus
\BIBentryALTinterwordstretchfactor\fontdimen3\font minus
  \fontdimen4\font\relax}
\providecommand{\BIBforeignlanguage}[2]{{%
\expandafter\ifx\csname l@#1\endcsname\relax
\typeout{** WARNING: IEEEtran.bst: No hyphenation pattern has been}%
\typeout{** loaded for the language `#1'. Using the pattern for}%
\typeout{** the default language instead.}%
\else
\language=\csname l@#1\endcsname
\fi
#2}}
\providecommand{\BIBdecl}{\relax}
\BIBdecl

\bibitem{danforth2009development}
D.~R. Danforth, M.~Procter, R.~Chen, M.~Johnson, and R.~Heller, ``Development
  of virtual patient simulations for medical education,'' \emph{Journal For
  Virtual Worlds Research}, vol.~2, no.~2, 2009.

\bibitem{danforth2013can}
D.~Danforth, A.~Price, K.~Maicher, D.~Post, B.~Liston, D.~Clinchot, C.~Ledford,
  D.~Way, and H.~Cronau, ``Can virtual standardized patients be used to assess
  communication skills in medical students,'' in \emph{Proceedings of the 17th
  Annual IAMSE Meeting, St. Andrews, Scotland}, 2013.

\bibitem{maicher2019using}
K.~R. Maicher, L.~Zimmerman, B.~Wilcox, B.~Liston, H.~Cronau, A.~Macerollo,
  L.~Jin, E.~Jaffe, M.~White, E.~Fosler-Lussier \emph{et~al.}, ``Using virtual
  standardized patients to accurately assess information gathering skills in
  medical students,'' \emph{Medical teacher}, vol.~41, no.~9, pp. 1053--1059,
  2019.

\bibitem{ruan2020towards}
W.~Ruan, Y.~Nechaev, L.~Chen, C.~Su, and I.~Kiss, ``Towards an asr error robust
  spoken language understanding system,'' \emph{Proc. Interspeech 2020}, pp.
  901--905, 2020.

\bibitem{wilcox2011chatscript}
``Chatscript,'' http://chatscript.sourceforge.net/, {Accessed}: December 2018.

\bibitem{jin2017combining}
L.~Jin, M.~White, E.~Jaffe, L.~Zimmerman, and D.~Danforth, ``Combining cnns and
  pattern matching for question interpretation in a virtual patient dialogue
  system,'' in \emph{Proceedings of the 12th Workshop on Innovative Use of NLP
  for Building Educational Applications}, 2017, pp. 11--21.

\bibitem{stiff2020self}
A.~Stiff, Q.~Song, and E.~Fosler-Lussier, ``How self-attention improves rare
  class performance in a question-answering dialogue agent,'' in
  \emph{Proceedings of the 21th Annual Meeting of the Special Interest Group on
  Discourse and Dialogue}, 2020, pp. 196--202.

\bibitem{sunder2020handling}
V.~Sunder and E.~Fosler-Lussier, ``Handling class imbalance in low-resource
  dialogue systems by combining few-shot classification and interpolation,''
  \emph{ICASSP}, 2021.

\bibitem{stiff2019improving}
A.~Stiff, P.~Serai, and E.~Fosler-Lussier, ``Improving human-computer
  interaction in low-resource settings with text-to-phonetic data
  augmentation,'' in \emph{ICASSP 2019-2019 IEEE International Conference on
  Acoustics, Speech and Signal Processing (ICASSP)}.\hskip 1em plus 0.5em minus
  0.4em\relax IEEE, 2019, pp. 7320--7324.

\bibitem{serai2022hallucination}
P.~Serai, V.~Sunder, and E.~Fosler-Lussier, ``Hallucination of speech
  recognition errors with sequence to sequence learning,'' \emph{IEEE/ACM
  Transactions on Audio, Speech, and Language Processing}, 2022.

\bibitem{namazifar2020warped}
M.~Namazifar, G.~Tur, and D.~H. T{\"u}r, ``Warped language models for noise
  robust language understanding,'' \emph{IEEE SLT 2021}, 2020.

\bibitem{gehring2017convolutional}
J.~Gehring, M.~Auli, D.~Grangier, D.~Yarats, and Y.~N. Dauphin, ``Convolutional
  sequence to sequence learning,'' in \emph{ICML-Volume 70}, 2017, pp.
  1243--1252.

\bibitem{cho2014learning}
K.~Cho, B.~van Merri{\"e}nboer, C.~Gulcehre, D.~Bahdanau, F.~Bougares,
  H.~Schwenk, and Y.~Bengio, ``Learning phrase representations using rnn
  encoder--decoder for statistical machine translation,'' in \emph{Proceedings
  of the 2014 Conference on Empirical Methods in Natural Language Processing
  (EMNLP)}, 2014, pp. 1724--1734.

\bibitem{lin2017structured}
Z.~Lin, M.~Feng, C.~N.~d. Santos, M.~Yu, B.~Xiang, B.~Zhou, and Y.~Bengio, ``A
  structured self-attentive sentence embedding,'' \emph{ICLR}, 2017.

\bibitem{pennington2014glove}
J.~Pennington, R.~Socher, and C.~D. Manning, ``Glove: Global vectors for word
  representation,'' in \emph{Proceedings of the 2014 conference on empirical
  methods in natural language processing (EMNLP)}, 2014, pp. 1532--1543.

\bibitem{johnson2019billion}
J.~Johnson, M.~Douze, and H.~J{\'e}gou, ``Billion-scale similarity search with
  gpus,'' \emph{IEEE Transactions on Big Data}, 2019.

\bibitem{serai2019improving}
P.~Serai, P.~Wang, and E.~Fosler-Lussier, ``Improving speech recognition error
  prediction for modern and off-the-shelf speech recognizers,'' in \emph{ICASSP
  2019-2019 IEEE International Conference on Acoustics, Speech and Signal
  Processing (ICASSP)}.\hskip 1em plus 0.5em minus 0.4em\relax IEEE, 2019, pp.
  7255--7259.

\bibitem{sutskever2013importance}
I.~Sutskever, J.~Martens, G.~Dahl, and G.~Hinton, ``On the importance of
  initialization and momentum in deep learning,'' in \emph{ICML}, 2013, pp.
  1139--1147.

\bibitem{kingma2014adam}
D.~P. Kingma and J.~Ba, ``Adam: A method for stochastic optimization,''
  \emph{ICLR}, 2015.

\bibitem{mani2020asr}
A.~Mani, S.~Palaskar, N.~V. Meripo, S.~Konam, and F.~Metze, ``Asr error
  correction and domain adaptation using machine translation,'' in \emph{ICASSP
  2020-2020 IEEE International Conference on Acoustics, Speech and Signal
  Processing (ICASSP)}.\hskip 1em plus 0.5em minus 0.4em\relax IEEE, 2020, pp.
  6344--6348.

\bibitem{kim2020two}
S.~Kim, G.~Kim, S.~Shin, and S.~Lee, ``Two-stage textual knowledge distillation
  to speech encoder for spoken language understanding,'' \emph{ICASSP}, 2021.

\end{thebibliography}


\end{document}